# Conditional Outlier Detection for Clinical Alerting

**Milos Hauskrecht, PhD[1], Michal Valko, MSc[1], Iyad Batal, MSc[1], Gilles Clermont, MD, MS[2], Shyam Visweswaran MD, PhD[3], Gregory F. Cooper, MD, PhD[3]**
**[1] Computer Science Department, [2] Department of Critical Care Medicine, [3] Department of Biomedical Informatics, University of Pittsburgh, Pittsburgh, PA, US**

**Abstract**

*We develop and evaluate a data-driven approach for detecting unusual (anomalous) patient-management actions using past patient cases stored in an electronic health record (EHR) system. Our hypothesis is that patient-management actions that are unusual with respect to past patients may be due to a potential error and that it is worthwhile to raise an alert if such a condition is encountered. We evaluate this hypothesis using data obtained from the electronic health records of 4,486 post-cardiac surgical patients. We base the evaluation on the opinions of a panel of experts. The results support that anomaly-based alerting can have reasonably low false alert rates and that stronger anomalies are correlated with higher alert rates.*

## Introduction

Despite numerous improvements in the health care practice, the occurrence of medical errors remains a persistent and serious problem [1, 2]. The urgency and the scope of this problem prompt the development of solutions aimed to aid clinicians in eliminating such mistakes. Current computer tools for monitoring patients are primarily knowledge-based; the ability to monitor depends on the knowledge represented in the computer and extracted a priori from clinical experts. Unfortunately, these systems are time consuming to build and their clinical coverage is not complete.

This paper describes a new monitoring and alerting framework we have developed that relies on stored clinical information of past patient cases and on statistical methods for the identification of clinical outliers (anomalies). Our conjecture is that the detection of anomalies corresponding to unusual patient management actions will help to identify medical errors. We believe that such an approach can complement the use of knowledge-based alerting systems, thereby improving overall clinical coverage of alerting. In clinical sub-areas where knowledge-based alerting is not yet available, this new approach can serve as a standalone system.

Typical anomaly detection attempts to identify unusual data instances that deviate from the majority of examples in the dataset [3]. Our objective is different; we want to identify anomalies in patient management actions (decisions), where individual actions depend strongly on the condition of the patient. We develop a new detection framework to identify outliers in such settings. Our framework builds upon the conditional outlier approach [4] that aims to identify patient management actions for a given patient that are highly unusual with respect to past patients and the condition he/she suffers from. Once a deviation is detected it may be used to generate a patient-specific alert.

To illustrate the idea, consider a post-surgical cardiac patient treated by heparin for the past six days whose platelet count fell below 100 and whose platelet count is decreasing. This patient may have heparin induced thrombocytopenia (HIT), a life threatening condition, which, if not treated properly, may lead to thrombosis and death. Assuming the platelet drop cannot be explained by other causes (e.g., hemorrhaging), a special HPF4 lab assay confirming HIT would typically be ordered. If ordering an HPF4 test is a typical action in such a situation, then seeing no HPF4 test order for such a patient is anomalous, and thus, it becomes the basis for raising an alert for consideration by the clinician(s) who are caring for the patient.

Ideally we would like to have all statistical outliers correspond to medical errors. Hence their detection would lead to a useful clinical alert. However, the reality is that a patient management action that is unusual from a statistical viewpoint does not always correspond to a helpful clinical alert. Nevertheless, our belief (and hypothesis) is that unusual patient management actions that can be found by anomaly-detection methods will flag errors often enough to be worth alerting on. We report here our investigation of the relationship between statistical outliers and clinically useful alerts. We conducted experiments on retrospective data obtained from electronic health records (EHRs) of 4,486 post-cardiac surgical patients and by having a subset of the 222 alerts raised by our system evaluated by a panel of 15 experts in critical care. We show that statistical anomaly detection is positively correlated with clinically meaningful alerts, which provides support for our hypothesis that anomaly-based alerting can be clinically useful.

## Methodology

This section presents a framework for identifying patient-management anomalies using data from past patients and for using them to generate patient specific alerts.

### Conditional anomaly detection

Let $\boldsymbol{x}$ define a vector of attributes (representing a patient's state) and $y$ a patient-management action that is associated with it. Our goal is to decide if the action $y$ is unusual for the patient condition $\boldsymbol{x}$, by taking into account records for past patients in the database. Our approach for indentifying conditional anomalies consists of three steps: (1) segmentation and transformation of temporal information in EHRs, (2) building of predictive models for different patient management actions, and (3) prospective, application of conditional anomaly detection to identify highly unusual actions in a given patient over the course of that patient's clinical care.

### Segmentation and transformation of the EHR

The data in patient medical records consist of complex multivariate time series combining results of lab tests, information about medication orders, procedures performed, diagnoses made, events encountered, and other information. In order to build an outlier detection model for patient-management actions, we first segment the data in every medical record according to discrete segmentation points and convert the time series data observed up to these segmentation points into a vector space representation of patient states at that time. Each patient-state instance is also linked to a vector of patient-management actions (decisions) made in between the current and the next segmentation point. This approach lets us generate multiple patient-state/action pairs per patient. Figure 1 illustrates the idea using 24-hour segmentation; the EHR for case A is segmented into three patient-state instances. Patient management actions are linked to the patient state features that preceded them.

### Vector space representation of the patient state

The segmentation divides the patient case into multiple patient-case instances. These instances cover different hospitalization periods and the amount of information in each of them may vary. To make our models independent of these variations, we use a vector space representation of the patient-state, where each state is defined using a fixed set of features and their values. The features attempt to summarize temporal information for each patient instance, and may include features such as last blood glucose measurement, last glucose trend value, and the total time the patient was most recently treated on heparin [4, 6].

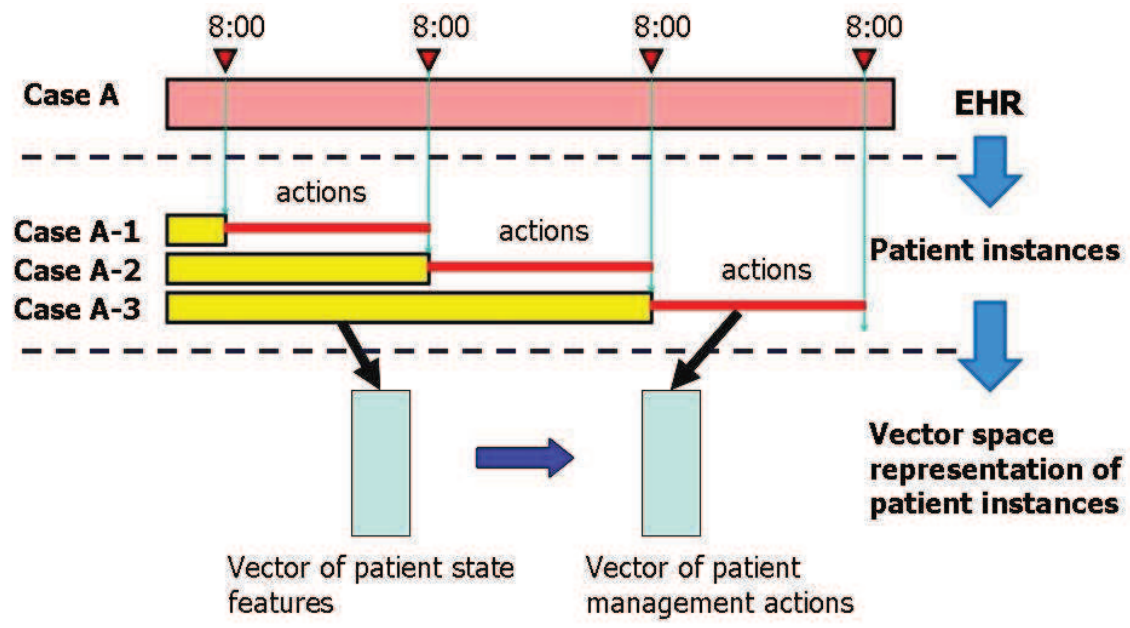

**Figure 1.** Processing of data in the electronic health record: (1) segmentation of an EHR into multiple patient-state/action instances, (2) transformation of these instances into a vector space representation of patient states and their follow-up actions.

The feature-based representation of time series data is flexible and various features can be built into the model. In this work we use three sets of features representing: laboratory tests (LABs) and their values, medications (MEDs), and procedures performed. We now briefly describe the features generated for each of these categories:

**Lab Features.** For labs with categorical results (e.g., POS/NEG) we used the following 7 features: last value; second to last value; first value; time since last value; a pair of indicators of the test being performed and pending orders. For labs with continuous or ordinal values we used a richer set of features, including features such as the difference between last two measured values, the slope in between the last two values, and their percentage drop/increase. Figure 2 illustrates a subset of features generated for such labs. In this case, the total number of features generated is 26 and it includes features for nadir, apex, baseline values and their differences from the last measured values. Nadir and apex values are the lab values with the smallest and the greatest value recorded up to that point, respectively.

**Medication Features.** For each medication we used four features: 1) indicator if the patient is currently on the medication, 2) time since the patient was put on that medication for the first time, 3) time since the patient was last on that medication, and 4) time since last change in the order of the medication.

**Procedure Features.** Procedure features capture information about procedures such as heart-valve repair. We record three features per procedure: 1) an indicator of whether the procedure has ever been performed during the current hospitalization, 2) the time since the procedure was first performed, and 3) the time since the procedure was last performed.

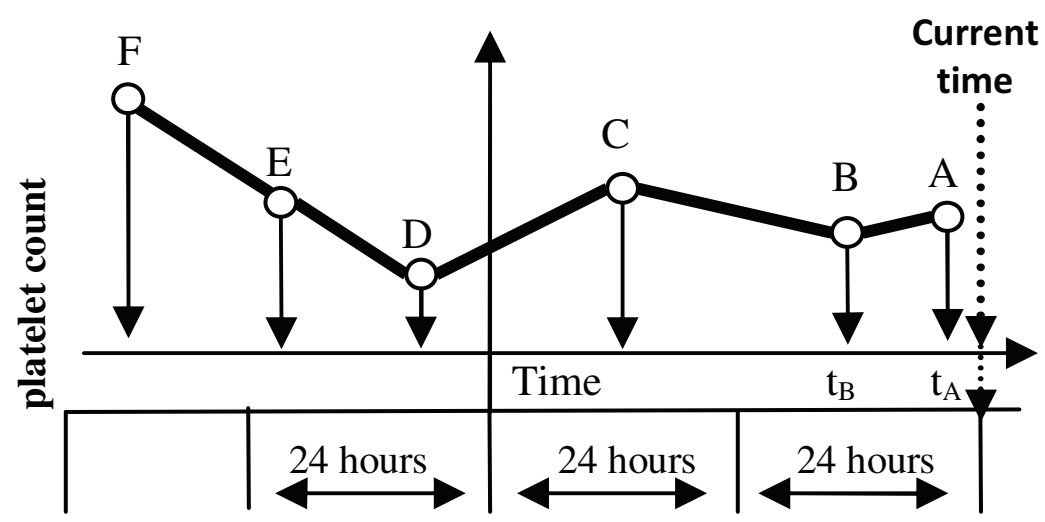

**Figure 2. Examples of temporal features for continuous lab values:** Last value: A**,** Last value difference = A-B**,** Last % change = (A-B)/B**,** Last slope = (A-B) / ($t_A$-$t_B$)**,** Nadir = D**,** Nadir difference = A-D, Nadir % difference = (A-D)/D.

### Representation of patient-management actions

We want to build predictive models that capture the relationship between the patient state and various follow-up patient-management actions. We consider two types of actions: (1) lab order actions with (true/false) values reflecting whether the lab was ordered or not, (2) medication actions with (true/false) values reflecting if the patient was given a medication or not. The values for all action are summarized in an action vector that is linked to the corresponding patient-state vector.

### Learning a predictive model

To assess the conditional anomaly of an *action y for a patient instance* $\boldsymbol{x}$*,* we first build (learn) a predictive model of data representing a conditional distribution $P(y \mid \boldsymbol{x})$. We can use such a model to determine if an actual action (or a planned action that is recorded) is unusual. Intuitively, the action *y is anomalous given* $\boldsymbol{x}$, if $P(y \mid \boldsymbol{x})$ is small.

We can construct the predictive model $P(y \mid \boldsymbol{x})$ using a variety of Machine Learning (ML) methods. In the work reported here we use the support vector machine (SVM) [7]. The SVM is a discriminative learning model that is popular in the machine learning community primarily thanks to its ability to learn high-quality discriminative patterns in high-dimensional datasets. It learns a decision boundary $f(\boldsymbol{x}) = 0$ that optimally separates examples into two classes. Hence, it does not directly output a probabilistic model of $P(y \mid \boldsymbol{x})$. However, multiple probabilistic transformations of the SVM model are possible [8]. We adopt the method by Platt in which

$$P(y=1 \mid \boldsymbol{x}) = 1/(1+\exp(A f(\boldsymbol{x}) + B)),$$

where A, B are parameters fitted on the training data.

### Feature selection

The total number of features in the vector-space patient-state representation is typically huge. It is not feasible to use all these features when learning predictive models. First, a particular action is likely to be influenced by only a relatively limited set of clinical variables (lab values, medications) while other clinical information is often irrelevant for that action. Second, if all features are incorporated into the classification model via parameters, then a high variance of these parameter estimates may negatively influence the quality of the model's predictions. To reduce this effect, a simpler model with a smaller number of features, and hence a smaller number of parameters, is often desirable. We approach the problem by building one model per action, and by selecting features for the model greedily and in groups, where groups are defined by different labs (e.g., hemoglobin features), medications (e.g., features for penicillin) or procedures (e.g., features linked to heart valve replacement). The quality of the features in the same group (e.g,. hemoglobin features) is determined by learning a multivariate model with these features only and by analyzing its predictive power via internal cross-validation in terms of the Area under the Receiver Operating characteristic (AUC) curve, which is equivalent to the *Wilcoxon-Mann-Whitney statistic* [9]. The groups are then combined incrementally into the model, by starting from the best group and greedily adding new feature groups into the model if they lead to the improvement of model's predictive performance. The process stops when no improvement is possible.

### Anomaly detection

Predictive models $P(y \mid \boldsymbol{x})$ built above capture how well patient instances predict different patient-management responses. The model can be applied to a (new) patient instance $\boldsymbol{x}$ and its associated action $y$ to see how different the actual action taken for the patient is from the predicted action.

**Anomaly scores.** We propose to measure the anomalousness of $y$ given $\boldsymbol{x}$ in terms of an anomaly score $Anom(\boldsymbol{x}, y)$. The most straightforward option is to define the score as $Anom(\boldsymbol{x}, y) = 1\text{-}P(y \mid x)$. Intuitively, the action *y is anomalous* given $\boldsymbol{x}$, if $P(y \mid \boldsymbol{x})$ is small or, equivalently, when $1 - P(y \mid \boldsymbol{x})$ is large. Other anomaly scores are possible, and we are in the process of exploring them as well.

### Outlier detection for clinical alerting

The above anomaly score allows us to measure deviations from usual patient-management patterns. Our hypothesis is that these deviations are often clinically important and may correspond to patient-management errors; hence they are worthwhile to be alerted on. But how do we use anomaly scores to alert on data encountered prospectively?

The caveat of applying our predictive models to detect anomalies prospectively is that all predictive models relate a patient state $\boldsymbol{x}$ with actions that follow it. Hence, a deviation in the patient management can be assessed only at the end of the follow-up period. But, by that time the patient state may change. To address the problem, we analyze each patient management action with respect to the patient state before and after the follow-up interval and assess their anomalousness. If the action is anomalous for both, then an alert is raised. We implement this idea in terms of the *alert score*. Let $\boldsymbol{x}_t$ be the patient instance at time $t$ (current time), $\boldsymbol{x}_{t-1}$ be the patient instance in the previous time step ($t$-1), and $y_{t-1}$ be the action regarding a lab order or a medication order in between times $t$-1 and $t$. We define the score for alerting on action $y_{t-1}$ given patient state $\boldsymbol{x}_t$ at time $t$ as:

$$Alert(x_t, y_{t-1}) = \min(Anom(x_{t-1}, y_{t-1},), Anom(x_t, y_{t-1},)).$$

Briefly, the score reflects: (1) the degree of deviation of the action $y_{t-1}$ as defined by the anomaly score, and (2) the persistence of a high anomaly signal in two consecutive time steps: ($t$-1) and $t$. If the score is above a threshold, an alert is raised at time $t$.

## Evaluation study

We evaluated our framework and its ability to raise clinically useful alerts using the data from archived EHRs for 4,486 post-surgical cardiac patients treated at a large teaching hospital in the Pittsburgh area. These EHRs were first divided into two groups: a training set that included all cases seen before 2005, and a test set that included the cases seen on or after 2005. Second, we used the time-stamped data in each EHR to segment the record at 8:00am every day to obtain multiple patient case instances, as illustrated in Figure 1. These patient instances were then converted into: (1) a vector-space representation of the patient state using the feature transformation described in the methodology section, and (2) a vector representation of lab-order and medication actions with true/false values, reflecting whether the lab was ordered or the medication was given within a 24-hour period. The segmentation led to 51,492 patient-state instances, such that 30,828 were used for training the model and 20,664 were used in the evaluation. The vector space representation of the patient state (at any point in time) included 9,282 generated features for 335 labs, 407 medications, and 36 procedure categories. Rare labs, medications and procedures (used in less than 20 patients) were excluded. Additional features we used were demographics (sex, age, race) and indicators of the presence of four heart-support devices. The action vectors linked with each patient-instance included 335 lab-order and 407 medication decisions.

**Learning anomaly detection models.** The training set was used to build three types of anomaly detection models: (1) models for detecting unexpected lab-order omissions, (2) models for detecting unexpected medication omissions, (3) models for detecting unexpected continuation of medications (commissions).

**Selection of alerts for the study.** The alerts for the evaluation study were selected as follows. We first applied all the above anomaly detection models to matching patient instances in the test set. The following criteria were then applied. First, only models with AUC of 0.68 or higher were considered. This means that many predictive models built did not qualify and were never used. Second, the minimum anomaly score for all alert candidates had to be 0.15 or lower. Third, for each action, only the strongest 125 anomalies and the strongest 20 alerts obtained from the test data were considered as alert candidates. This lead to 3,768 alert candidates, from which we selected 222 alerts for 100 patients, such that 101 alerts were lab-omission alerts, 55 were medication-omission alerts, and were 66 medication-commission alerts. Figure 3 shows the distribution of alerts in the study according to the alert score.

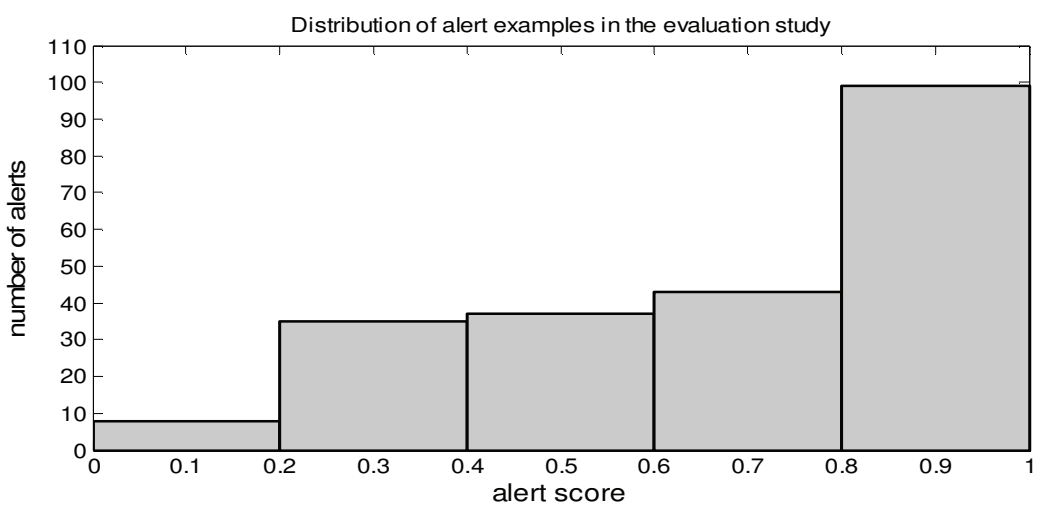

**Figure 3.** Histogram of alert examples in the study according to their alert score.

**Alert reviews.** The alerts selected for the study were assessed by physicians with expertise in post-cardiac surgical care. The reviewers (1) were given the patient cases and model-generated alerts for some of the patient management actions, and (2) were asked to assess the clinical usefulness of these alerts. We recruited 15 physicians to participate in the study, of which 12 were fellows and 3 were faculty from the Departments of Critical Care Medicine or Surgery. The reviewers were divided randomly into five groups, with three reviewers per group, for the total of 15 reviewers. Overall, each clinician made assessments of 44 or 45 alerts generated on 20 different patients. The survey was conducted over the internet using a secure web-based interface [10].

**Alert assessments.** The pairwise kappa agreement scores for the groups of three ranged from 0.32 to 0.56. We used the majority rule to define the gold standard. That is, an alert was considered to be useful if at least two out of three reviewers found it to be useful. Briefly, out of 222 alerts selected for the evaluation study, 121 alerts were agreed upon by the panel (via the majority rules) as useful alert.

**Analysis of clinical usefulness of alerts.** We analyzed the extent to which the alert score from a model was predictive of it producing clinically important alerts. Figure 4 summarizes the results by binning the alert scores (in intervals of the width of 0.2) and presenting the true alert rate per bin. The true alert rates vary from 19% for the low alert scores to 72% for high alert scores, indicating that higher alert scores are indicative of higher true alert rates. This is also confirmed by (1) a positive slope of the line in Figure 4, which was obtained by fitting the results via linear regression, and (2) the results of the ROC analysis. Briefly, all alerts reviewed were ordered according to their alert scores, from which we generated an ROC curve. The AUC for our alert score was 0.64. This is statistically significantly different from 0.5, which is the value one expects to see for random or non-informative orderings. Again, this supports that higher alert scores induce better true alert rates. Finally, we would like to note that alert rates in Figure 4 are promising and despite alert selection restrictions, they compare favorably to alert rates of existing clinical alert systems [10, 11].

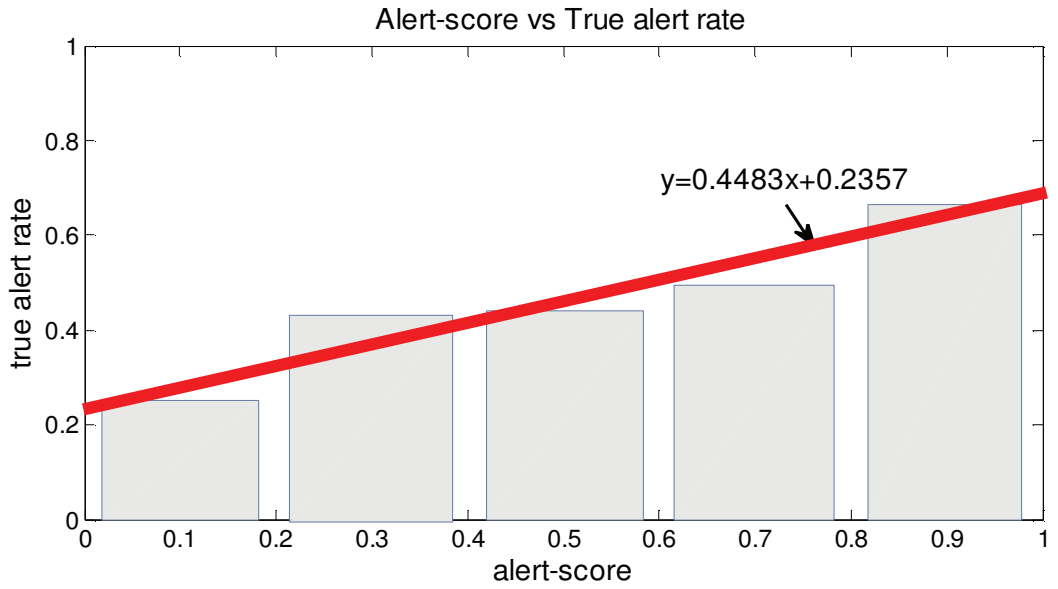

**Figure 4.** The relation between the alert score and the true alert rate. The height of the bins shows true alert rates for alert-score intervals of width 0.2. The line is fitted via linear regression.

## Conclusions

We presented a new framework that detects conditional anomalies in patient management actions and raises an alert when an anomaly is found. This work differs from previous studies we reported in [4,5,12], by building and testing the outlier detection models on thousands of patient features and hundreds of actions extracted from EHRs, and by building alerting models prospectively monitoring the patient case in time. It is the most comprehensive and complete evaluation of our conditional outlier methodology for detecting unusual patient management decisions we have conducted so far.

Our results show that outlier detection is a promising methodology for raising clinically useful alerts. In the future we plan to further investigate the framework by eliminating some of the restrictions placed on the selection of alerts in the study, and by using (1) new temporal features that characterize the clinical data, (2) different time discretizations, (3) relations in between actions, and (4) additional evaluations that contain different datasets in different clinical domains.

## Acknowledgements

We would like to thank Drs. Andrew Post and James Harrison for their PROTEMPA case review interface. This research work was supported by grants R21LM009102, R01LM010019, and R01GM088224 from the NIH. Its content is solely the responsibility of the authors and does not necessarily represent the official views of the NIH.